\documentclass{article}

    \usepackage[preprint]{neurips_2024}

\usepackage{graphicx}
\usepackage[utf8]{inputenc} % allow utf-8 input
\usepackage[T1]{fontenc}    % use 8-bit T1 fonts
\usepackage{hyperref}       % hyperlinks
\usepackage{url}            % simple URL typesetting
\usepackage{booktabs}       % professional-quality tables
\usepackage{amsfonts}       % blackboard math symbols
\usepackage{nicefrac}       % compact symbols for 1/2, etc.
\usepackage{microtype}      % microtypography
\usepackage{xcolor}         % colors
\usepackage{amsmath}

\title{ESQA: Event Sequences Question Answering}

\author{%
  Irina Abdullaeva\\
  AIRI \\
  Moscow, Russia\\
  \And
  Andrei Filatov \\
  Sber AI, Skoltech \\
  Moscow, Russia \\
  \AND
  Mikhail Orlov \\
  Sber AI Lab \\
  Moscow, Russia \\
  \And
  Ivan Karpukhin \\
  Sber AI Lab \\
  Moscow, Russia \\
  \And
  Viacheslav Vasilev \\
  Sber AI, MIPT \\
  Moscow, Russia \\
  \And
  Denis Dimitrov \\
  Sber AI, AIRI \\
  Moscow, Russia \\
  \And
  Andrey Kuznetsov \\
  AIRI,  Sber AI \\
  Moscow, Russia \\
  \And
  Ivan Kireev \\
  Sber AI Lab \\
  Moscow, Russia \\
  \And
  Andrey Savchenko \\
  Sber AI Lab \\
  Moscow, Russia \\
}

\begin{document}
\graphicspath{{figures/}}

\maketitle
% \footnotetext[1]{Artificial Intelligence Research Institute}

\begin{abstract}

% ORIGINAL.
% In the realm of \textbf{event sequences}, unlike in computer vision or natural language processing, leveraging a pre-trained model for solving multiple problems and adapting to new ones is uncommon. Current methods face limitations in flexibility, generalization, and computational efficiency. Moreover, integrating lengthy event sequences into neural network-based models remains challenging.

% To overcome these obstacles, we introduce a novel approach called \textbf{E}vent \textbf{S}equences \textbf{Q}uestion \textbf{A}nswering (\textbf{ESQA}), leveraging Large Language Models (LLM). We represent event sequence tasks in a question-answer format and propose a universal encoding method using a Transformer-based trainable encoder. Efficient feature extraction and significant sequence length reduction are achieved through the Q-Former model, acting as a connector layer between the encoder and LLM. Our empirical findings highlight ESQA's ability to rival state-of-the-art methods across various prediction tasks in multi-task scenarios using diverse financial datasets. Furthermore, ESQA exhibits adaptability to new tasks, outperforming statistical baselines.

% By Ivan
Event sequences (ESs) arise in many practical domains including finance, retail, social networks, and healthcare. In the context of machine learning, event sequences can be seen as a special type of tabular data with annotated timestamps. Despite the importance of ESs modeling and analysis, little effort was made in adapting large language models (LLMs) to the ESs domain. In this paper, we highlight the common difficulties of ESs processing and propose a novel solution capable of solving multiple downstream tasks with little or no finetuning. In particular, we solve the problem of working with long sequences and improve time and numeric features processing. The resulting method, called ESQA, effectively utilizes the power of LLMs and, according to extensive experiments, achieves state-of-the-art results in the ESs domain.

\end{abstract}

\section{Introduction}

% ORIGINAL.
% Temporal data, which is crucial in various fields like geoscience~\citep{Karianne2019geoscience}, healthcare~\citep{Esteva2019healthcare}, sociology~\citep{Hossain2020Prediction}, industry~\citep{Choi2021Detection}, e-commerce~\citep{ni2018perceive} and finance~\citep{babaev2022coles}, often comes in the form of complex event sequences. An event sequence is defined as a set of chronologically ordered events that are connected by cause or time. Unlike a regular time series, each event in an event sequence combines characteristics from multiple sources, making them distinct in nature.

% By Ivan
Temporal data often comes in the form of event sequences, where each event is characterized by the arrival time and additional structured data. This type of data is widely spread in domains like geoscience~\citep{Karianne2019geoscience}, healthcare~\citep{Esteva2019healthcare}, sociology~\citep{Hossain2020Prediction}, industry~\citep{Choi2021Detection}, e-commerce~\citep{ni2018perceive} and finance~\citep{babaev2022coles}. Event sequences combine properties of time series and tabular data while having major differences. Unlike time series, events can arrive with irregular time steps and can have structured annotations, similar to tabular datasets. Unlike tabular data, events have timestamps and associated order. These differences require special data processing, modeling, and inference approaches.

% ORIGINAL.
% Exploring and handling time-based data is a new frontier in deep learning, yet existing methods often focus narrowly on specific challenges, lacking the versatility and deep understanding needed to address diverse tasks simultaneously and adapt to new ones seamlessly.

% Cutting-edge methods for analyzing structured time-based data use language models, transforming tabular data into text~\citep{hegselmann2023tabllm, jin2023time}. Yet, for instance, a client's transaction history can contain extensive details about their behavior over time. The sheer volume of entries, coupled with the complex attention mechanism will inevitably lead to performance degradation.

% By Ivan.
The new frontier in machine learning, especially in deep learning, focuses on adapting large language models (LLMs) to domains beyond language. The reasons behind this adaptation is that LLMs can use additional information, not found in domain-specific data, can process textual context of the underlying task, generate answers in a free natural form, can argue its decisions and support dialog with the user. The potential benefits of using LLMs include improved modeling quality and generalization. The latter means that the hybrid model can solve new problems with little or no finetuning, that largely increases the applicability of the model and reduces development costs. Successful applications of LLMs were demonstrated in both time series \citep{cai2023jolt} and tabular datasets \citep{dinh2022lift}, but no effort was made to adapt LLMs to event sequences: financial transactions, electronic health records, activity on different devices and so on. These data characterise a human live and are used to personalise many AI services across different domains.

Event sequence processing with LLM encounters several difficulties. First, structured data must be effectively encoded at the LLM's input. Textual representation considerably increases the sequence length and can't be effectively processed by modern Transformer models due to the quadratic complexity. Second, the desired method must be capable of processing long input sequences, even when the downstream tasks require historical data analysis. The problem is similar to the first but focuses on the model architecture rather than input processing. Finally, time feature and the order must be properly provided to the model, as they constitute the essence of event sequences and include important information for solving downstream tasks.

In this paper, we propose a new neural architecture, called ESQA, that exploits the power of LLMs to model event sequences and to solve associated practical tasks. In particular, we for the first time develop a question-answering approach with LLM backbone in the event sequences domain. We show the proposed model is capable of solving multiple downstream tasks without finetuning. When finetuned, ESQA outperforms other methods and achieves a new state-of-the-art.

\section{Background}
\textbf{Event Sequences.} We assume that events, denoted as $e_i$, are arranged in sequences $S_n = {\{e_i\}}_{i=1}^{I_n}$ based on their association with a common entity. Here, ${I_n}$ represents the number of events in the sequence $S_n$. An entity could represent a bank customer or a web user, while the events within the sequence might include actions like a completed transaction or a series of clicks. These events are connected by a temporal order: $t(e_i) < t(e_{i+1})$, where $t(.)$ indicates the time at which the event occurs.
Event sequences encompass a diverse range of attributes, with each event, $e_i$, characterized by a set of features ${\{c_j\}}_{j=1}^C$. These features can be depicted as a vector of values with dimension $C$. Additionally, $Y_m$ represents the target variable vector for the problem at hand, which may be based on the value of a sequence feature $c_m$ or external variables, such as a bank client's default status. Attributes of events comprise both numeric $c_j^{num}$ and categorical features $c_j^{cat}$ of various types. Categorical features define attribute values within a finite set of categories $c_j^{cat} \in |c_j| = \{cat_{j;1}, ..., cat_{j;K_j}\}$, where $K_j$ denotes the number of possible values for the feature $c_j^{cat}$~\citep{lane2003introduction}. Numerical features $c_j^{num} \in \mathbb{R}$ are those represented as numbers, allowing meaningful arithmetic operations to be performed~\citep{lane2003introduction}.

\textbf{LLMs for Tabular Data.}
Large Language Models (LLMs) are a family of neural architectures pretrained on a large corpus of texts. LLMs accept inputs in the form of text and generate textual output. In practice, LLM architecture is composed of three main blocks. The first one is an embedding layer, that converts input text to a sequence of numeric vectors known as embeddings. The second block, the backbone, transforms input embeddings to the output embeddings sequence with possibly different length. The final part of the model maps embeddings to the output text.

There are two main approaches for encoding tabular data at the input of LLM. The first one is to provide a description of each table field in the textual form~\citep{dinh2022lift}. This approach suffers from little flexibility and extremely long input sequences. The second approach is to replace the embedding layer, with a newly designed module capable of directly encoding table fields to embeddings with the required number of features. The latter approach is also known as embedding injection and usually achieves better results~\citep{Koh2023GroundingLM,Huang2023LanguageIN}.

\textbf{Question Answering with LLMs.}
The popular way to solve problems with LLMs is to design a question such that a valid answer to this question solves the problem~\citep{dinh2022lift}. The question must include the context, i.e. all necessary data required for reasoning, and the task definition. This way LLM input is usually composed of the context, task, and connecting words indicating the boundaries of each part.

% \section{Event Sequences Question Answering (NEW)}
% The general view of the proposed model, called {\it Event Sequences Question Answering (ESQA)}, is presented in Figure \ref{fig:arch}. The main questions with LLM adaption include input coding and output processing. In this section we provide a detailed answer to these two questions.

% \subsection{Input Coding}
% ESQA implements embeddings injection at the LLM input to achieve effective coding of event sequences. The input sequence includes three main parts: static prompt, context, and the task-dependent question. Static prompt is trivial and is generated by a standard LLM embedding layer from the text "This is client's transactions history", enclosed by two special tokens: [AMNT] and [TRX]. The task-dependent question is also generated by a standard embedding layer, but the text depends on the particular downstream task. The context coding requires more clarification and is described in the subsections below.

% \subsection{Parameter Tuning}
% \textbf{Pretraining for zero-shot prediction.}

% \textbf{Finetuning.}

%\section{Proposed Approach}
\section{Event Sequences Question Answering}
\begin{figure*}
  \centering
  \includegraphics[width=0.9\linewidth]{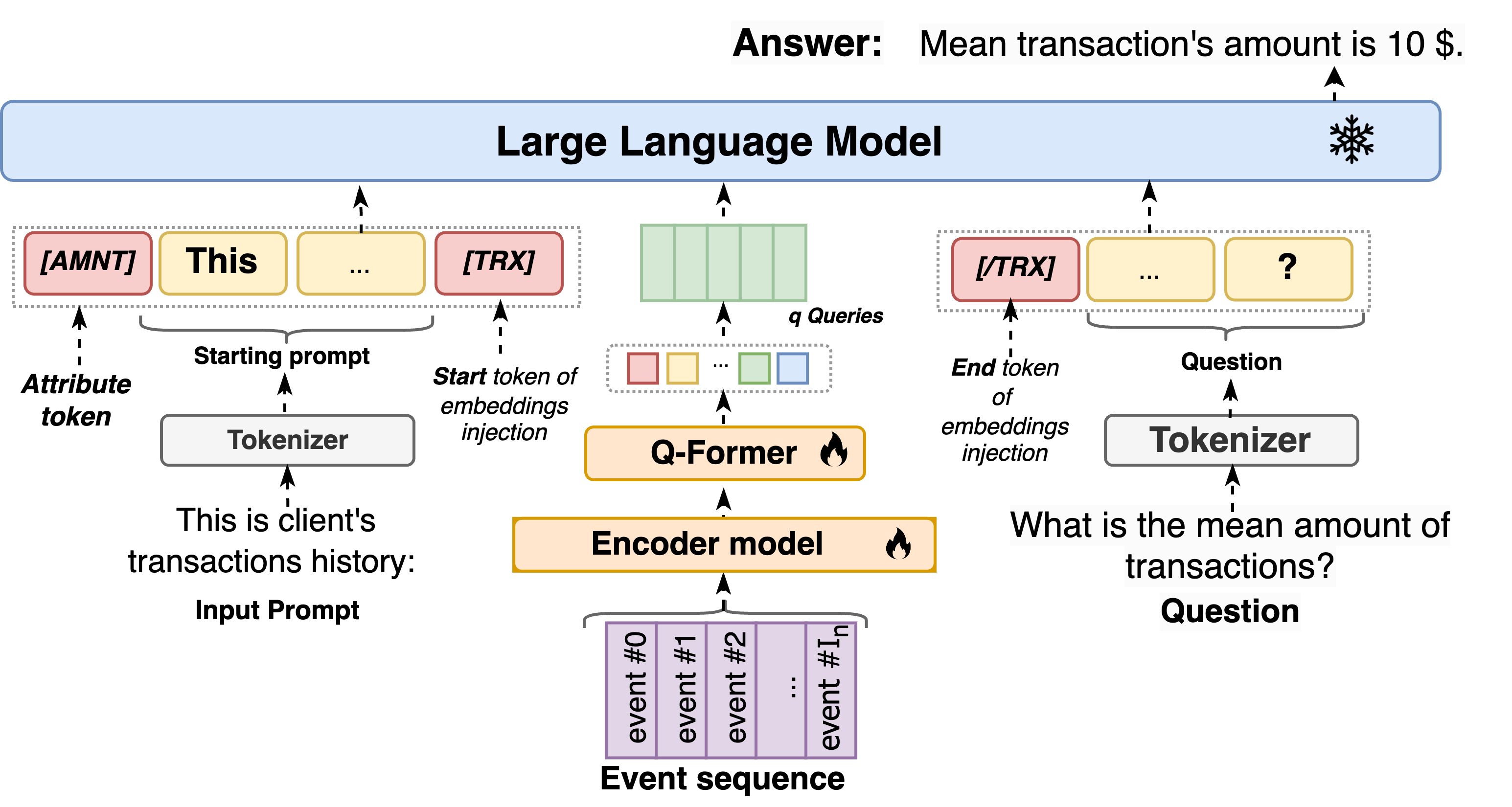}
  \caption{Model architecture. The components of the approach that do not require training are colored in blue. Components whose weights are optimised during training are colored in orange. The trainable embeddings and associated tokens are colored in red.}
  \label{fig:arch}
\end{figure*}

%The overall architecture of the approach and its components are summarized in Fig.~\ref{fig:arch}.
The general view of the proposed model, called {\it Event Sequences Question Answering (ESQA)}, is presented in Figure \ref{fig:arch}. Below we will give a detailed description of the model's input and the backbone LLM.

\subsection{Questions and answers construction}

% Multiple downstream tasks solving requires a way to task definition and transfer to the model. We use the power of LLM and assume a task definition in a form of natural language question. 

The concept behind this method is to frame all tasks involving temporally structured data as natural language questions and answers. Each task from ${\{task_m\}}_{m=1}^M$ takes the form $\{Q_m, X_m, A_m\}$, where $Q_m$ is the question that defines the problem, $X$ represents the input data, and $A_m$ is the answer sought based on the target variable $Y_m$. 

% This structured task representation is depicted in Figure 2a.

A question $Q_m$ consists of two components: the prefix and the question body. The prefix initiates the query token sequence and is placed before embeddings of other modalities. The question body then describes the task in textual form. For example, the task of determining the most frequent value of feature $c_m$ is represented as: \textit{``What is the most frequent value of $c_m$ in the entire dataset?''}.

Given the nature of time-structured data, we classify questions into two types: extractive and predictive. Extractive questions focus on tasks involving existing event sequences, such as computing statistics or identifying trends and characteristics. Predictive questions, on the other hand, pertain to tasks concerning the prediction of future events or attributes based on available data.

Tasks and their corresponding questions can also be categorized based on the type of response sought: binary, multiple choice, or open-ended. Binary questions seek a straightforward answer, either as $A_m \in \{0, 1\}$ or in the form of \textit{``Yes''} or \textit{``No''}. For instance, a question like \textit{``Is drinking water the most frequently purchased product?''} can be answered with a simple \textit{``Yes''} or \textit{``No''}.

In contrast to binary questions, multiple-choice and open-ended questions assume a specific answer corresponding to the required feature, whether numerical $A_m \in \mathbb{R}$ or categorical $A_m \in |c_j| = \{cat_1, \dots, cat_K\}$. Multiple-choice questions provide a list of possible answer choices. For example, one might ask \textit{``What is the most frequently purchased product? Options: black tea; bread; drinking water; grapes.''}. Open-ended questions, on the other hand, prompt a direct response, such as \textit{``What is the name of the most frequently purchased product? Please provide the name in your response.''}.

%\subsection{Event sequences modality injection}
%\subsection{Event sequences injection}
\subsection{Events embeddings}
To address the integration of event sequences into a language model, we propose adapting the method outlined in previous works~\citep{Koh2023GroundingLM,Huang2023LanguageIN}. This involves embedding multi-modal information into an LLM, parameterized by ${\theta}$, by directly mapping it into the intrinsic embedding space $E^{\theta}$, bypassing the discrete text token layer. To achieve this, we introduce a trainable mapping $\phi: Z \rightarrow E^{\theta}$, where $Z$ represents the observation space of temporally structured data. This mapping converts the data into a sequence of $f$-dimensional vectors in$E^{\theta}$, which are then integrated into a sequence of text embeddings. This interleaving of modalities creates a multi-modal input for the LLM.

%\subsection{Model}

%\subsection{Event sequences encoding}

%Our approach proposes representing all event features as trainable embeddings. To achieve this, we first encode each 
ESQA represents all event features as trainable embeddings. It is achieved by encoding each value $x_{ij}$ of a categorical or integer numeric feature $c_j$ with a sequential index $k_{x_{ij}}$ based on the total number of unique values for that feature $k = [0, \dots, K_j]$. This index uniquely identifies the embedding $emb_k$ of a feature value in the embedding matrix $W_e$. The embedding dimension is selected based on the formula: $dim(e_k) = \lceil{\lambda \times K_j^{\mu}}\rceil$. The coefficients $\lambda = 1.6$ and $\mu = 0.56$ have been chosen empirically.

Numerical features in the form of real numbers are discretized into non-overlapping intervals: $B_j^1, \dots, B_j^n$, $B_j^i = [b_j^{i-1}, b_j^i) $. The distribution of the feature $c_j$ in the training sample is used to determine these intervals. The number of intervals is chosen based on the approach in~\citep{Doane1976AestheticFC}, using the formula $n = 1 + \log_2 (n) + \log_2 (1 + \frac{|g_1|}{\sigma_{g_1}})$, where $g_1$ is the estimated third-moment skewness of the distribution and $\sigma_{g_1} = \sqrt{\frac{6(n-2)}{(n+1)(n+3)}}$. This method is particularly suited for distributions of features that deviate significantly from the normal distribution.

Once the intervals have been defined, Eq.~\ref{eq:1} determines the value $x_{j;disc}^{num}$ of the $j$'th numerical feature:

\[
\label{eq:1}
x_{j;disc}^{num} = 
    \begin{cases}
        b_j^0, & x_{ij} < b_j^0, \\
        b_j^n, & x_{ij} \geq b_j^n, \\ 
        b_j^i & b_j^{i-1} \leq x_{ij} < b_j^i.
    \end{cases}
\]

\begin{figure}[h]
  \centering
  \includegraphics[width=0.9\linewidth]{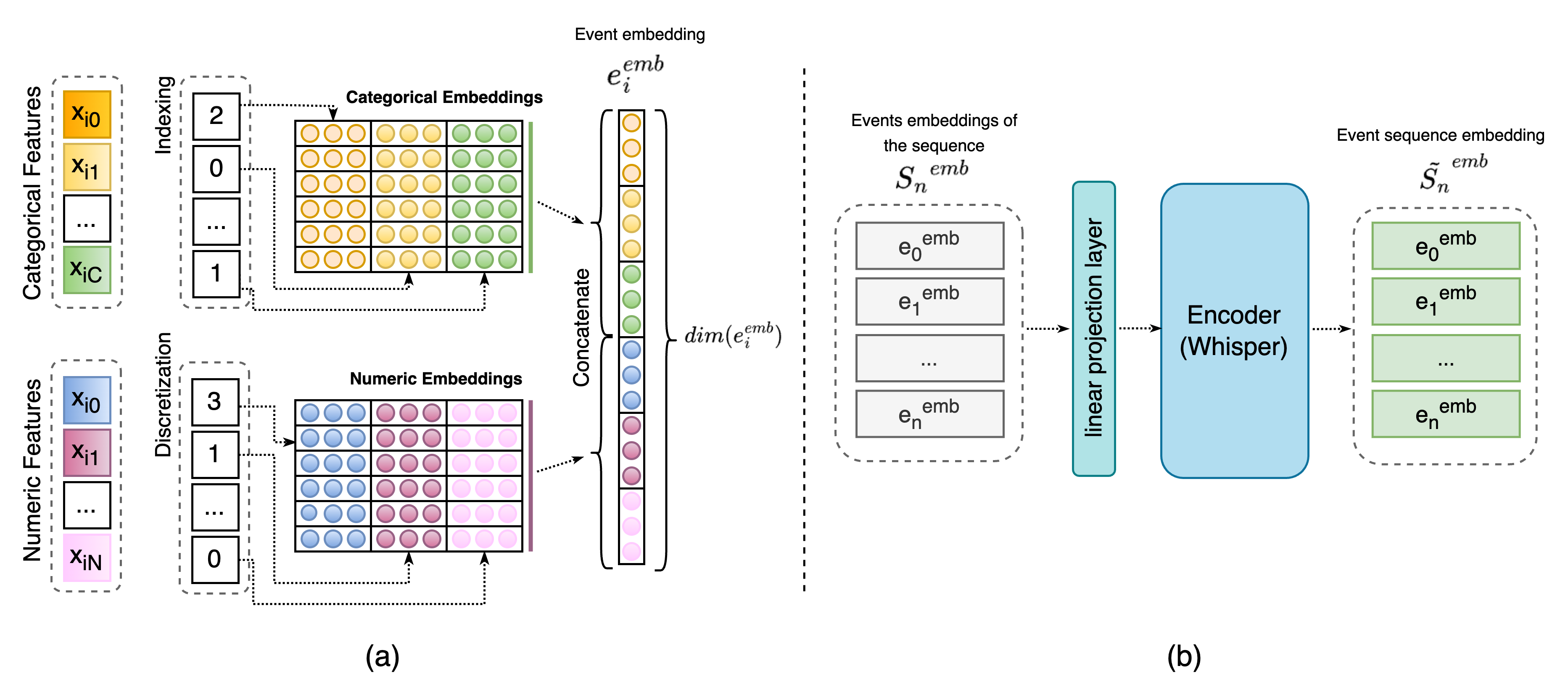}
  \caption{a) Event sequences features encoding; in the example, there are $N$ numerical and $C$ categorical features, which are concatenated into a tensor $e_i^{emb}$ of dimension $dim(e_i^{emb})$. b) The event sequence encoder model processes the concatenated feature embedding vectors $S_n^{emb}$ for all events within a sequence, ultimately producing a comprehensive embedding $\tilde{S_n}^{emb}$ for the entire event sequence.}
  \label{fig:fproc}
\end{figure}

The resulting feature embeddings are concatenated into a tensor $e_i^{emb}$ of dimension $dim(e_i^{emb}) = \sum_{j=1}^{C} |c_j|$, which describes a single event $e_i$ from the sequence. A vector representation of sequence $S_n$ is formed by combining vector representations of individual events into a joint tensor $S_n^{emb}$ shown on Fig.\ref{fig:fproc}a.

%In addition to the approaches for dealing with numerical and temporal features, we conducted a number of experiments with other encoding techniques, the results of which are summarised in the Appendix \ref{}.

\subsection{Encoder}
After the initial layer of input data embeddings, vectorized event sequences are fed into a specialized encoder model Fig.~\ref{fig:fproc}b. This module, based on the architecture of the Transformer decoder, processes sequences of events in an autoregressive manner by predicting each subsequent event. For our implementation, we used both Whisper-tiny and Whisper-small models~\citep{Radford2022RobustSR}, initialized with weights pre-trained on audio data. The input tensor for the encoder comprises concatenated feature embedding vectors for all events $S_n^{emb}$ (Section 3.3.1) and has a size of $dim(S_n^{emb}) = (I_n, dim(e_i^{emb}))$. The encoder processes this tensor autoregressively, similar to the sequence of text token embeddings, resulting in a sequence of vectors $\tilde{S_n}^{emb}$ with a size $dim(\tilde{S_n}^{emb}) = (I_n, d_{enc})$. Here, $d_{enc}$ represents the output layer dimensionality of the encoder model. To ensure compatibility between the dimensions of the input embeddings of the event sequences $dim(S_n^{emb})$ and the embedding layer of the encoder model $d_{enc}$, we used a linear projection layer.

This choice of encoder architecture is motivated both by the temporal nature of the event sequences, which aligns with autoregressive modelling, and by the results of a series of experiments. Appendix~\ref{A.1} provides a detailed description of the experiments and their results.

\subsection{Connector}

\begin{figure}[h]
  \centering
  \includegraphics[width=0.95\linewidth]{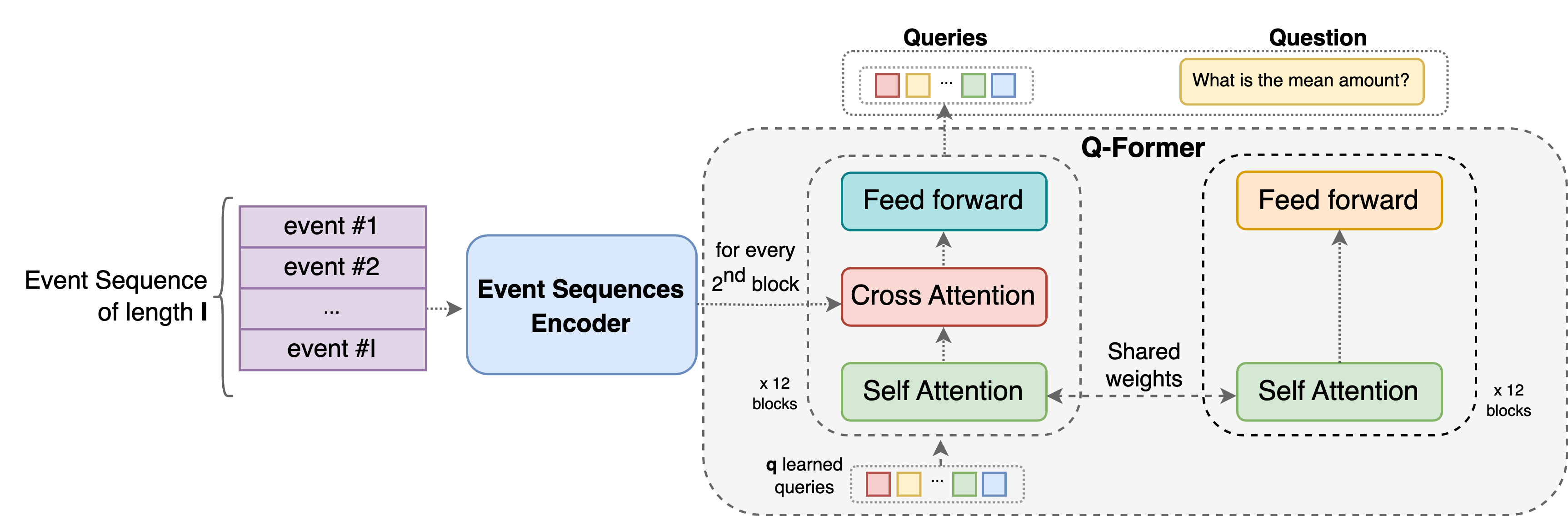}
  \caption{The Q-Former model's architecture is designed to extract the most relevant event sequence representations. It produces $q$ query embeddings for each event sequence, which are then linearly projected to the size of the language model embedding and appended to the embedded question tokens. Subsequently, the joint sequence is transmitted to the LLM.}
  \label{fig:qformer}
\end{figure}

The output representation of the event sequence encoder grows in dimensionality as the number of events in each sequence increases. This size is crucial, as it must fit within a common multimodal embedding sequence, impacting the extension of the language model's context length. Our goals are to shorten the event sequence length without significant information loss and to adapt each event's vectorized representation to match the language model's embedding dimension. To achieve this, we propose an intermediate connection layer between the event sequence encoder and the LLM. We suggest using the Query Transformer model, or Q-Former~\citep{Li2023BLIP2BL}, to efficiently extract features from the encoder output.

The Q-Former architecture (Fig.~\ref{fig:qformer}) includes two transformer submodules: a novel modality transformer (originally an image transformer) that works with a fixed image encoder for feature extraction, and a text transformer that functions as both an encoder and a decoder. A set of trainable query embeddings $q$ serves as the input for the novel modality transformer. These queries engage in self-attention, interacting with each other and with the fixed modality features through cross-attention layers in every other transformer block. 

In our approach, Q-Former produces $q$ query vectors for each event sequence, which are then passed to the LLM. We use a single fully-connected layer to project the output query vectors into the language model's text embedding dimension. In this study, we initialize Q-Former with the weights from the BLIP-2 approach, derived from BLIP-2 with the FLAN-T5-xl model~\citep{Li2023BLIP2BL}. The architecture and initialization of the connection layer were chosen based on a series of experiments detailed in Appendix~\ref{A.2}.

\subsection{Language Model}
As the backbone for the pre-trained LLM, our approach utilizes the FLAN-T5 family of encoder-decoder models \citep{Wei2021FinetunedLM}. Any process of fine-tuning model parameters influences the model's proficiency in a specific domain but also causes it to "forget" essential general and linguistic knowledge. To preserve this knowledge and save computational resources, we have frozen most of the LLM parameters. Studies \citep{Lu2021PretrainedTA, Zhou2023OneFA} indicate that freezing most of the model's weights often yields better results than fully fine-tuning a pre-trained LLM.

To efficiently select a limited set of trainable parameters, we propose using Parameter-Efficient Fine-Tuning (PEFT) methods. Specifically, we employed the Low-Rank Adaptation (LoRA) approach \citep{Hu2021LoRALA}, which keeps most of the model weights frozen while adding trainable rank decomposition matrices to a subset of the parameters.

%The overall architecture of the approach and its components are summarized in Fig.~\ref{fig:arch}.

\section{Experiments}

In this section, we begin by presenting the evaluation details, which include the comparison methods and the datasets used for evaluation. Following this, we conduct a series of systematic experiments to showcase the capability of the developed ESQA approach in addressing a diverse range of problems based on event sequences.

\subsection{Experimental setup}

\textbf{Datasets.} \label{sec:datasets} Sequences of events are prevalent across various domains and tasks, with a particularly high demand for analyzing such data in the fintech sector. In this field, transactional activity of individuals serves as the primary source of information. Consequently, we have chosen to utilize a collection of datasets containing customer transactions from banks and marketplaces as examples of event sequences. The sensitivity of the information in these datasets has significantly influenced our choice and the number of datasets used, given the limited availability of public datasets in this area.

We selected five publicly available datasets with event sequences. These include: AlfaBattle2.0 \citep{Smirnov2021AlfaBattle2.0}, Age Group Prediction Competition \citep{Sirius2020sberbank-sirius-lesson}, X5 Retail Hero: Uplift Modelling for Promotional Campaign from \citep{babaev2022coles}, Taiwan Default of Credit Card Customers from \citep{Yeh2009TheCO}, and Gender Prediction Competition from \citep{MaxValery2019python-and-analyze-data-final-project}. The data were divided into two subsets, training and validation sets, ensuring no overlap by unique client identifiers. The proportions for partitioning these sets were chosen independently for each dataset. A detailed description of the datasets used to evaluate the approach's quality is provided in Appendix \ref{B.3}.

\textbf{Baselines.} We have chosen representative baseline approaches for analysing event sequences, which have demonstrated effectiveness across various benchmarks. For the AlfaBattle and Taiwan Default of Credit Card Clients datasets, we implemented, trained, and fine-tuned the baseline models ourselves to achieve optimal results. For other benchmarks, we relied on the findings from CoLES \citep{babaev2022coles}, which provide the most current and comprehensive empirical studies of event sequences. 

For the next event prediction task we also provide calculated statistical baselines for both numerical and categorical target features to compare the quality of prediction tasks in zero-shot setups. These include, for example, the prediction of the mean or median value for numerical target variables and the prediction of the most frequent value for categorical attributes. 

A complete list and detailed description of baselines is provided in Appendix \ref{B.4}.

\subsection{Experimental results} \label{sec:results}

\subsubsection{Main results}

Each dataset used for model evaluation corresponds to a specific downstream task. For instance, the AlfaBattle dataset is utilized for predicting a bank customer's loan default, while the Age dataset is employed for predicting the age group. It is important to highlight that the AlfaBattle dataset is highly imbalanced, with the positive class constituting less than 3\%, while the Gender dataset has a slight over-representation of the class denoting male gender. Therefore, we used the ROC-AUC metric for problems with binary target variables and class imbalance. For multiclass classification with balanced classes, we employed Accuracy. A more detailed explanation of the metric calculation methodology and the assessment of response quality for ESQA is provided in Appendix~\ref{B.2}. The results of the experiments on the downstream tasks for datasets described in Section~\ref{sec:datasets} are summarized in Table~\ref{tab:main_res}.

\begin{table}
  \caption{A comparison of ESQA on the downstream tasks of the five event sequence datasets described in Section 4.1 with the baseline approaches presented in Section 4.2.1. The best results are highlighted in \textbf{bold} and the second best results are \underline{underlined}.}
  \label{tab:main_res}
    \centering
    \begin{tabular}{lccccc} 
    \toprule
        \textbf{Dataset} & \textbf{AlfaBattle} & \textbf{Age} & \textbf{Gender} & \textbf{X5} & \textbf{Taiwan} \\
        Metric & AUCROC & Accuracy & AUCROC & Accuracy & AUCROC \\
        \midrule
        Handcrafted feat. & 0.7792 & 0.629 & \underline{0.877}& \underline{0.547}&  \underline{0.784}\\
        Randomly init. RNN & 0.6456 & 0.375 & 0.593 &   -& 0.722 \\
        CPC & \underline{0.7919} & 0.602 & 0.851 & 0.525 & - \\
        Barlow Twins & 0.7878 & 0.634 & 0.865 &  -& -  \\
        CoLES & \textbf{0.7921}& \underline{0.640}& \textbf{0.881}& 0.539 &  -  \\
        NSP & 0.7655 & 0.621 & 0.852 & 0.425 &  -  \\
        RTD & 0.7910 & 0.631 & 0.855 & 0.520 &  -  \\
        SOP & 0.7238 & 0.512 & 0.785 & 0.428 &  - \\
        MLM NSP & 0.7591 &  -&  -&  -&  -   \\
        TabFormer & 0.7862 &  -&  -&  -&  -  \\
        GPT & 0.7737 &  -&  -&  -&  -  \\ 
        \hline
        \textbf{ESQA (ours)} &  0.7568 & \textbf{0.699} & 0.850 & \textbf{0.598} & \textbf{0.793}\\
    \bottomrule
    \end{tabular}
\end{table}

The results indicate that the ESQA approach matches both self-supervised contrastive and supervised methods in quality. Notably, on the Age and X5 datasets, ESQA surpasses the baseline scores. Although specific comparative results for other models are not available for the Taiwan dataset, ESQA's impressive performance underscores its effectiveness. These outcomes highlight ESQA's superior capability in handling multi-class classification tasks with balanced classes.

However, for the client default problem on the AlfaBattle dataset and the Gender dataset, the results are less clear-cut. The CoLES contrastive approach achieves the highest quality for these problems. While ESQA slightly lags behind CoLES, it still shows competitive performance, closely following models like Barlow Twins and RTD, and outperforming the SOP approach. It is important to note that both datasets exhibit class imbalance, which is especially pronounced in the AlfaBattle case. 

This leads us to conclude that the ESQA approach generally performs classification tasks as well as, or better than, the selected baseline methods. However, it is significantly affected by the imbalance of the target variable. This limitation can be attributed to the nature of LLMs, originally designed to extract common patterns from text data to model complex language structures.

\subsubsection{Predictive tasks}

The majority of tasks involving event sequences require answering predictive questions about event features. To address such challenges, we propose utilizing the ESQA approach in a multi-task setting, enabling simultaneous predictions of all features of the next event in the sequence. Experimental results for predictive questions against baselines are detailed in Table~\ref{tab:pred_res}.

\begin{table}
    \caption{Table comparing ESQA with the baseline approaches presented in Section 4.2.1 for predicting attributes of the next transaction on the AlfaBattle dataset. The best results are highlighted in \textbf{bold} and the second best results are \underline{underlined}.}
    \label{tab:pred_res}
    \centering
    \begin{tabular}{lccc} 
        \toprule
        \textbf{Attribute} & \textbf{MCC code} & \textbf{Amount} & \textbf{Hour diff} \\
        Metric & Acc./F1 & MAE/MSE & MAE/MSE \\
        \midrule
        CoLES & 0.440 / 0.351 & 0.197 / 0.082 & 36.05 / 1586.52 \\
        CPC & 0.475 / 0.411 & 0.196 / 0.074 & 34.89 / 1508.71 \\
        RNN with CoLES & 0.469 /0.411 & 0.184 /0.077 & 32.25 / 1573.02\\
        CatBoost & 0.440 /0.367 & 0.190 /0.090 & 34.40 / 1613.41 \\
        GPT with descr. & 0.462 /0.423 & 0.179 / 0.083 & 32.63 / 1726.42 \\
        Text LLM  & 0.382 / 0.381 & \textbf{0.103 / 0.0176}~ & 116.38 / 62161 \\ 
        \hline
        ESQA (ours) & \textbf{0.546 / 0.546} & 0.191 / 0.1021 & \textbf{18.313 / 1033.87} \\
        \bottomrule
    \end{tabular}
\end{table}

On categorical feature prediction tasks, such as MCC code attribute prediction, ESQA achieves the highest performance with an Accuracy/F1 scores, outperforming all other models, with the closest being CPC. This indicates that ESQA is particularly effective in handling categorical prediction tasks within the transaction history context. 

In predicting the numerical Amount attribute, while the Text LLM achieves the lowest MAE/MSE, ESQA still performs competitively. Although ESQA is not the top performer here, it maintains reasonable accuracy, demonstrating its versatility across different types of prediction tasks.

For the temporal Hour diff attribute, ESQA significantly outperforms all other models. The next best model, RNN CoLES, has a much higher MAE/MSE, highlighting ESQA's superior capability in handling temporal prediction tasks effectively.

\subsubsection{Generalization abilities}

LLMs possess an extraordinary capacity to generalise to novel, previously unseen tasks. Our method maintains the integrity of the language model's weights, thereby preserving its inherent capabilities. Moreover, by training adaptors within the attention layers, we expand the domain of zero-shot tasks from exclusively text-based tasks to those based on event sequences. We evaluated the ESQA approach's adaptability to new predictive tasks following comprehensive pre-training on a set of contextual tasks. The model was trained in  multi-task setting on all event features of the AlfaBattle dataset and was subsequently tested in a zero-shot setting across various predictive tasks within the same dataset. Table~\ref{tab:zeroshot_res} presents a comparison of our experimental results against statistical baselines, a text baseline, and an ESQA model specifically trained on those predictive tasks.

\begin{table}
    \caption{Table comparing the generalisation abilities of the ESQA approach with the statistical baseline approaches presented in Section 4.2.1, and a text-based approach. The ESQA approach trained on predictive tasks in a multitask setting is referred to as 'ESQA m/t'. While ESQA trained on contextual tasks and adapting to new tasks is referred to as 'ESQA z/s'. The best results are highlighted in \textbf{bold} and the second best results are \underline{underlined}.}
    \label{tab:zeroshot_res}
    \centering
    \begin{tabular}{llcccc} 
        \toprule
        \textbf{Attribute} & \textbf{Stat. baseline} & \textbf{Text-only} & \textbf{ESQA m/t} & \textbf{ESQA z/s} \\
        MCC code, acc. & \underline{0.388} & 0.382 & \textbf{0.546} & 0.381 \\
        MCC category, acc. &  0.437& 0.402 & \textbf{0.588} & 0.435 \\
        Amount, MAE/MSE & 0.241 & \textbf{0.103/0.018} & \underline{0.191/0.102} & 0.389/0.228 \\
        City, acc. &   \underline{0.704}& 0.691& \textbf{0.731} & 0.343 \\
        Country, acc. &  0.970& 0.970 & \textbf{0.972} & \underline{0.971} \\
        Currency, acc. & 0.987 & 0.986~ & \underline{0.987}~ & \textbf{0.988~} \\
        Op. type gr., acc. &  0.766& 0.733 & \textbf{0.840} & \underline{0.781} \\
        Op. type, acc. &  0.499& 0.393 & \textbf{0.633} & \underline{0.543} \\
        Op. kind, acc. &  0.548& 0.494 & \textbf{0.693} & \underline{0.598} \\
        Days before, MAE/MSE  &  140.5 / 23823.3& \underline{10.5 / 657.2} & \textbf{6.3/195.9} & 11.394 / 666.2 \\
        Hour diff, MAE/MSE & \underline{36.33} & 116.4/62161 & \textbf{18.3/1033.9} & 48.85/3980 \\
        \bottomrule
\end{tabular}
\end{table}

For the MCC code and MCC category attributes, ESQA multi-task outperforms all baselines, indicating its strength in handling categorical predictions. However, in the zero-shot setting, ESQA's performance is comparable to the statistical baseline, suggesting room for improvement in scenarios without task-specific training. In predicting the Amount attribute, the text-only approach achieves the best MAE/MSE, while ESQA multi-task shows competitive performance, demonstrating its robustness in handling regression tasks despite not being the top performer. However, the regression problem on real numbers with a large number of decimals is still a challenging task for zero-shot ESQA, which performed poorly. For temporal predictions like Days before and Hour diff, ESQA multi-task significantly outperforms other approaches, showcasing its superior capability in modelling temporal patterns. Overall, ESQA zero-shot performance, while not leading, still provides valuable insights into ESQA's versatility and potential for improvement in less customised settings.

\section{Related work}
%In recent times, natural language processing has seen significant progress with the rise of LLMs~\citep{Achiam2023GPT4TR, Raffel2019ExploringTL, Touvron2023LLaMAOA,Jiang2023Mistral7}. Beyond text, these models have shown significant progress in processing diverse data types like images~\citep{Koh2023GroundingLM, Liu2023VisualIT, Huang2023LanguageIN, Zhang2023LLaMAAdapterEF,Gao2023LLaMAAdapterVP} and audio~\citep{Han2023ImageBindLLMMI}. In this section, we explore how language models are used in fields akin to event sequences and the specific models designed for handling such sequences.

\textbf{Event sequences.}
Temporal Point Processes (TPPs) and their marked variants (MTPPs) can be seen as the simplest forms of event sequences. Previous research was focused on accurate next event prediction with or without neural networks~\citep{liniger2009multivariatehawkes,mei2017neuralhawkes,xue2023easytpp}. %In this work, we consider a complex event structure with multiple fields, including categorical and numeric features. 
Another branch of research addressed event streams of the general form \cite{padhi2021tabular,babaev2022coles,mcdermott2024eventstreamgpt}. To the best of our knowledge, question answering with LLMs was not previously applied to TPPs nor event sequence modeling.

\textbf{Structured modeling with neural networks.} The problem of modeling multiple heterogeneous features with neural networks was addressed in tabular neural networks~\citep{Yin2020TaBERTPF, Iida2021TABBIEPR, padhi2021tabular, Hegselmann2022TabLLMFC, Yang2022TableFormerRT, dinh2022lift}. We reuse best practices for making embeddings from categorical and numeric features. At the same time, event sequences require analysis of multiple events at once, while tabular datasets can be processed one row at a time. To this end, ESQA applies encoding method and adapt Q-Former, not seen in tabular neural networks.

\textbf{LLMs for time series.} Previous works made use of LLMs in the context of time series analysis~\citep{gruver2023large,cai2023jolt,zhang2024llmtssurvey}. Unlike time series models, ESQA implements a novel context encoding and can process complex data structures.

\section{Conclusion}

In this paper, we introduced Event Sequences Question Answering (ESQA), a novel approach for modelling event sequences with LLMs. %By adapting question answering techniques to the event sequence domain, we successfully applied ESQA to a wide array of diverse tasks.
Our empirical results demonstrate that our approach performs robustly across various datasets. For several downstream problems, ESQA performs at least as well as specialised baselines \ref{tab:main_res}, and for the task of predicting the attributes of the next event, it significantly surpasses baseline methods \ref{tab:pred_res}. Furthermore, we have shown that ESQA can handle multiple tasks simultaneously without any special fine-tuning \ref{tab:zeroshot_res}, highlighting its remarkable ability to adapt swiftly to new tasks without the need for complex and time-consuming training. These findings position ESQA as an exceptionally promising approach %first step in 
leveraging the strong generalisation capabilities of LLM backbones for the field of event sequences.

\textbf{Limitations.} \label{sec:limit} Our research has certain limitations. In processing numerical features, ESQA employs value discretization, which introduces an inherent discretization error. This error is significantly influenced by the number of discretization buckets and the ranges of the actual feature values. To mitigate this error, we conducted several additional experiments to refine the pre-processing method. Furthermore, handling time features in event sequences requires special attention. We are actively exploring ways to enhance temporal feature processing within ESQA. In future work, we will focus on implementing these improvements and addressing the challenge of dealing with unbalanced classes.

\clearpage
{
\small
\bibliographystyle{plainnat}
\bibliography{esqa}
}

%%%%%%%%%%%%%%%%%%%%%%%%%%%%%%%%%%%%%%%%%%%%%%%%%%%%%%%%%%%%

\newpage
\appendix
\onecolumn

\section{Architecture components selection}

\subsection{Event sequence encoder architecture selection} \label{A.1}

Training a language model to understand a different modality is not a novel challenge and has been addressed for various data types. Therefore, in our experiments on the event sequence encoder architecture, we built upon advancements from other modalities. We focused on established models for three highly developed modalities: text, images, and audio. For text architectures, we examined several models including encoder-only models like BERT (base, large), encoder-decoder models like T5 (small, base, large), and decoder-only models like GPT (base, medium, large). For image architectures, we used ViT (base, large). For audio models, we considered various versions of Whisper (tiny, small, medium), utilizing only the decoder part of the Whisper architecture.

The models were compared based on their ability to predict the default of a bank client in the AlfaBattle 2.0 dataset, a binary problem where the task is to determine if a bank client will repay a loan based on their transaction history over two years. We used AUC as the metric for comparison. All models were trained from scratch.

We maintained a consistent training scheme across all experiments, employing Adam as the optimizer with a learning rate of 1e-4. A linear warm-up of the learning rate was applied for the first epoch, followed by a linear decay to zero after 10 epochs. To ensure compatibility between the dimensions of transaction embeddings and the dimensions of pretrained model embeddings, we used a linear layer for text and audio models. Since ViT models cannot process sequences, we addressed this issue by applying a single layer of cross-attention to a fixed number of learnable latent tokens.

As shown in Table \ref{tab:enc_choice}, decoder-only models outperformed both encoder-only and encoder-decoder models in event sequence encoding in almost all setups. Specifically, experiments with text models demonstrated that the decoder-only GPT2 model outperformed the encoder-decoder T5 model, and the BERT model training did not converge. Similarly, audio architectures, which are primarily decoder-based, also showed superior performance. In response to the concerns about the performance of larger models within the same family, as observed in Table \ref{tab:enc_choice}, our analysis suggests that the enlargement of the encoder size contributes to overfitting. This overfitting is the primary reason for the degradation in performance outcomes.

\begin{table}[h]
    \caption{Table comparing different architectures for predicting default of the client on the AlfaBattle dataset. The best results are highlighted in \textbf{bold} and the second best results are \underline{underlined}.}
    \label{tab:enc_choice}
    \centering
    \begin{tabular}{lccc} 
        \toprule
        \textbf{Architecture} & \textbf{Type} & \textbf{Number of parameters} & \textbf{AUC} \\
        \midrule
        GPT2 Base      & Decoder         & 124M & 0.7869      \\
        GPT2 Medium    & Decoder         & 355M & 0.7833      \\
        GPT2 Large     & Decoder         & 774M & 0.7747      \\
        Whisper-tiny   & Decoder         & 29M  & \underline{0.7892}      \\
        Whisper-small  & Decoder         & 153M & 
        \textbf{0.7894}      \\
        Whisper-medium & Decoder         & 456M & 0.7715      \\ \hline
        T5 Small       & Encoder-Decoder & 60M  & 0.7721      \\
        T5 Base        & Encoder-Decoder & 223M & 0.7756      \\
        T5 Large       & Encoder-Decoder & 770M & Diverged           \\ \hline
        BERT Base      & Encoder         & 110M & Diverged \\
        BERT Large     & Encoder         & 335M & Diverged \\
        ViT Base       & Encoder         & 85M  & 0.7822      \\
        ViT Large      & Encoder         & 302M & 0.7639      \\
        \bottomrule
    \end{tabular}
\end{table}

After determining the type of architecture (i.e., the decoder), we conducted further experiments to identify the specific type and size of the decoder architecture. We compared Whisper-tiny, Whisper-small, and GPT2-base, as they produced the best results. Additionally, we evaluated various types and sizes of recurrent architectures: GRU-1, GRU-6, GRU-12, LSTM-1, and LSTM-4, where the number indicates the number of layers used in each model. The embedding size for all recurrent models was set to 1024.

\begin{table}[h]
    \caption{Table comparing different \textbf{decoder} architectures presented in Section 
    \ref{A.1} for predicting default and attributes of the next transaction on the AlfaBattle dataset. The best results are highlighted in \textbf{bold} and the second best results are \underline{underlined}.}
    \label{tab:final_enc_choice}
    \centering

\begin{tabular}{llllll} 
    \toprule
    \textbf{Architecture} & \textbf{\# params.} & \textbf{Amount} & \textbf{MCC Category} & \textbf{24-hour acc} & \textbf{Default} \\
    Metric & & MSE & Accuracy & Accuracy & AUC \\
    \midrule
    Whisper-tiny & 29 M. & 0.0660 & 0.4861 & Diverged & \underline{0.7892} \\
    Whisper-small & 153 M. & \textbf{0.0656} & \textbf{0.4896} & \underline{0.645} & \textbf{0.7894} \\
    GPT-2-base & 100 M. & \underline{0.0657} & \underline{0.4888} & Diverged & 0.7869 \\
    GRU-1 & 0.3 M. & 0.0668 & 0.4817 & 0.418 & 0.7854 \\
    GRU-big & 16 M. & 0.0670 & 0.4805 & Diverged & 0.7578 \\
    GRU-large & 35 M. & 0.0662 & 0.4815 & Diverged & 0.7732 \\
    LSTM-1 & 0.4 M. & 0.0669 & 0.4830 & 0.634 & 0.7710 \\
    LSTM-4 & 2 M. & 0.0664 & 0.4858 & \textbf{0.655} & 0.7664 \\
    \bottomrule
\end{tabular}

    % \begin{tabular}{lllll} 
    %     \toprule
    %     \textbf{Architecture} & \textbf{\# params.} & \textbf{MCC code} & \textbf{MCC category} & \textbf{Amount} \\
    %     Metric &  & Accuracy & Accuracy & MSE \\
    %     \midrule
    %     Linear & 197 k. & 0.501 & 0.574 & 0.0174 \\
    %     2 x Linear & 920 k. & 0.523 & 0.561 & 0.0196 \\
    %     RNN (LSTM) & 3.94 M. & 0.509 & 0.558 & 0.0220 \\
    %     Transformer & 1.1 M. & 0.478 & 0.529 & 0.1361 \\
    %     2 x Transformer & ??? M. & 0.519 & 0.555 & \underline{0.0168} \\
    %     Q-Former-small & 14.7 M. & 0.519 & \textbf{0.579} & \textbf{0.0162} \\
    %     Q-Former-base (w/o init.) & 96 M. & \underline{0.526} & \underline{0.570} & 0.0189 \\
    %     Q-Former-base (w. init.) & 96 M. & \textbf{0.527} & 0.569 & 0.0177 \\
    %     \bottomrule
    % \end{tabular}
    
\end{table}

% \begin{tabular}{lccccc}
%         \hline
%         \multicolumn{1}{c}{\textbf{Method}} & \textbf{Parameters} & \multicolumn{4}{c}{\textbf{Metric}}                                           \\ \hline
%         \multicolumn{1}{c}{}                &                     & Amount           & MCC             & 24-hour acc             & Default         \\ \hline
%         Whisper-tiny               & 29M                 & 0.066            & 0.4861          & Diverged                       & \underline{0.7892}    \\
%         Whisper-small              & 153M                & \textbf{0.06567} & \textbf{0.4896} & \underline{0.6452}            & \textbf{0.7894} \\
%         GPT-2-base                 & 100M                & \underline{0.06571}    & \underline{ 0.4888}    & Diverged                       & 0.7869          \\
%         GRU-1                      & 0.3M                & 0.06683          & 0.4817          & 0.4189                  & 0.7854          \\
%         GRU-big                    & 16M                 & 0.067            & 0.4805          & Diverged                       & 0.7578          \\
%         GRU-large                  & 35M                 & 0.06621          & 0.4815          & Diverged                       & 0.7732          \\
%         LSTM-1                     & 0.4M                & 0.06696          & 0.483           & \textit{0.6343}         & 0.771           \\
%         LSTM-4                     & 2M                  & 0.06644          & 0.4858          & \textbf{0.655} & 0.7664          \\ \hline
%     \end{tabular}

Table \ref{tab:final_enc_choice} indicates that transformer architectures outperformed recurrent models. Scaling up recurrent models did not significantly enhance their quality and sometimes even degraded their performance. Given the similar results among transformer architectures, we selected Whisper-small as the optimal model for all ESQA experiments.

\subsection{Connector architecture selection} \label{A.2}

\begin{table}
    \caption{Table comparing different \textbf{connector} architectures for better modalities alignment. Q-Former architecture based connectors with initialisations from BLIP-2 \citep{Li2023BLIP2BL} pretrained weights are labelled ‘w. init.’, without initialisation are indicated by ‘w/o. init.’. The best results are highlighted in \textbf{bold} and the second best results are \underline{underlined}.}
    \label{tab:conn_choice}
    \centering
    \begin{tabular}{lllll} 
        \toprule
        \textbf{Architecture} & \textbf{\# params.} & \textbf{MCC code} & \textbf{MCC category} & \textbf{Amount} \\
        Metric &  & Accuracy & Accuracy & MSE \\
        \midrule
        Linear & 197 k. & 0.501 & 0.574 & 0.0174 \\
        2 x Linear & 920 k. & 0.523 & 0.561 & 0.0196 \\
        RNN (LSTM) & 3.94 M. & 0.509 & 0.558 & 0.0220 \\
        Transformer & 1.1 M. & 0.478 & 0.529 & 0.1361 \\
        2 x Transformer & 2.09 M. & 0.519 & 0.555 & \underline{0.0168} \\
        Q-Former-small & 14.7 M. & 0.519 & \textbf{0.579} & \textbf{0.0162} \\
        Q-Former-base (w/o init.) & 96 M. & \underline{0.526} & \underline{0.570} & 0.0189 \\
        Q-Former-base (w. init.) & 96 M. & \textbf{0.527} & 0.569 & 0.0177 \\
        \bottomrule
    \end{tabular}
\end{table}

Integrating multiple modalities within a single approach centered around an LLM requires mapping new modalities into a textual model. Employing a separate encoder for each modality simplifies the task to finding an efficient architecture for mapping each modality's vector space to the LLM embedding text space. When analyzing event sequences, processing extended data sequences presents challenges due to increased context length, which leads to higher computational complexity. In some instances, the sequence length may surpass the maximum context length of the language model.

To address these challenges, we conducted experiments to determine the optimal architecture for the connection layer between the event sequence encoder and the LLM. We evaluated several potential implementations: a single linear layer, a transformer layer, and two model sizes of the Q-Former architecture. Additionally, we investigated the impact of initialization on problem-solving quality and training speed by initializing the Q-Former with weights from the pre-trained visual-text model BLIP-2, based on FLAN-T5. In all experiments, we tackled three tasks in a multi-task mode using the AlfaBattle dataset. The components used in all experiments included Whisper-tiny as the transaction encoder and FLAN-T5-small as the language model. Performance was measured at 20 epochs, with fixed batch size, learning rate, and optimization parameters. We used multi-class accuracy for classification tasks and MSE for numerical response prediction tasks as target metrics.

The results revealed that simply increasing the number of trainable parameters does not necessarily enhance task solution quality. A linear layer with a small number of parameters performed worse than Q-Former-small, which also trained much faster. However, adding more simple identical blocks within a single connector, such as \('2 x Linear'\), did not significantly improve performance. On the other hand, more complex blocks, such as \('2 x Transformer'\), showed substantial quality gains. Increasing the model size to Q-Former-base yielded mixed results: while MCC code prediction quality improved by 2\%, the metrics for MCC category prediction and numerical attribute Amount declined.

Additional initialization with weights from visual-text pre-training marginally improved the MCC code prediction task but slightly degraded the metrics for the other two tasks. The overall impact of initialization was minimal, indicating few common patterns between extracting salient information from images and deriving dependencies from event sequences. This discrepancy is expected due to the lack of temporal dependence within a single image, in contrast to the strong temporal dependence between events in a sequence.

Therefore, we selected the Q-Former-base model without initialization, anticipating an increase in the number of tasks our approach can handle simultaneously. This model offers a sufficient margin for increasing the complexity of future experiments.

\section{Implementation details}

\subsubsection{Training and hyper-parameters} \label{B.1}

All experiments for ESQA described below utilised consistent hyperparameters and approach components, unless otherwise specified. We employed the AdamW optimizer \citep{Loshchilov2017FixingWD} with parameters $\beta_1 = 0.9$, $\beta_2 = 0.98$, and a weight decay of 0.01. Cosine learning rate decay with restarts was applied, featuring different peak learning rates for each dataset and varying numbers of warm-up steps. In our experiments, LoRA \citep{Hu2021LoRALA} with a rank of $r=16$ was applied only to the matrices $W_q$ and $W_v$ of the self-attention and encoder-decoder attention layers. The LoRA scaling factor was set to 32, and the dropout rate to 0.05. The number of trainable parameters in the language model was calculated as $\theta^{train} = 2 \times L \times d_{model} \times r$, where $L$ is the number of layers and $d_{model}$ is the internal dimensionality of the language model. The rank of trainable decomposition matrices is denoted by $r$. Therefore, the number of trainable parameters in each FLAN-T5 model did not exceed 0.9\% of the total parameters (Table~\ref{tab:peft}). All models were trained using 6 Nvidia A100 (80G) GPUs. The training hyperparameters are summarised in Table~\ref{tab:hparams}.

 \begin{table}
  \caption{Hyperparameters used for ESQA training. In all experiments, the Whisper-small model architecture was used as the encoder.}
  \label{tab:hparams}
    \centering
    \begin{tabular}{lllllll} 
        \toprule
        Dataset & AlfaBattle2 & Age & Gender & X5 & Taiwan &  \\
        LLM & flan-T5-xl & FLAN-T5-xl & fla-T5-large & FLAN-T5-xl & FLAN-T5-xl &  \\
        Emb. size & 201 & 110 & 74 & 163 & 100 &  \\
        Learn. rate & 3e-4 & 3e-4 & 1e-4 & 1e-4 & 1e-4 &  \\
        warmup steps & 4 k. & 1 k. & 1 k. & 4 k. & 1 k. &  \\
        Max. epochs & 40 & 10 & 10 & 20 & 30 &  \\
        Batch size & 300 & 250 & 50 & 250 & 50 &  \\
        Min seq. len. & 50 & 0 & 0 & 0 & 6 &  \\
        Max seq. len. & 750 & 1500 & 1500 & 750 & 6 &  \\
        \bottomrule
    \end{tabular}
\end{table}

\begin{table}
  \caption{Trainable parameters of LLM with LoRA.}
  \label{tab:peft}
  \centering
  \begin{tabular}{lll}
    \toprule
    \cmidrule(r){1-3}
    Model     & \% trainable params.  & \# trainable params. \\
    \midrule
    FLAN-T5-small & 0.8862 & 0.688 M.     \\
    FLAN-T5-base  & 0.7096 & 1.77 M.      \\
    FLAN-T5-large & 0.5989 & 4.72 M.  \\
    FLAN-T5-xl & 0.3301 & 9.44 M.  \\
    FLAN-T5-xxl & 0.1692 & 18.87 M.  \\
    \bottomrule
  \end{tabular}
\end{table}

\subsubsection{Evaluation strategy} \label{B.2}

We employed several classical machine learning metrics to thoroughly evaluate the proposed approach. As previously mentioned, ESQA is designed to handle tasks that can be framed as binary or multi-class classification as well as regression settings.

\textbf{Classification Metrics.} We utilised classification metrics for tasks that involved predicting a categorical feature of the next event or a characteristic of the entire sequence (e.g., default of a bank customer). For non-binary target tasks, we used Accuracy and F1-score. For binary target tasks, we employed the Area Under the Receiver Operating Characteristic curve (ROC-AUC). The model with the highest performance on these metrics was deemed the best.

To calculate the classification metrics Accuracy and F1 score using the language model's response in the question-answer format, we applied the following process. The question body was followed by an instruction specifying the format of the answer to clearly define the structure of the language model's output. The tokens predicted by the language model were then decoded into text, and the segments containing the desired answer were extracted. These extracted values $y$ were compared to the target $\hat{y}$ in a classification format, where the number of classes matched the cardinality of the predicted value. Subsequently, Accuracy and F1 were calculated as follows:
\[
\label{eq:2}
\texttt{Accuracy}(y, \hat{y}) = \frac{1}{n_\text{samples}} \sum_{i=0}^{n_\text{samples}-1} 1(\hat{y}_i = y_i)
\]

\[
\label{eq:3}
\text{F1} = \frac{2 * \text{TP}}{2 * \text{TP} + \text{FP} + \text{FN}}
\]

In this context, $TP$ represents the number of true positives, $FP$ stands for the number of false negatives and $FP$ denotes the number of false positives. 

In calculating the ROC-AUC metric, we utilised the difference between the probabilities of the positive and negative response tokens.

\textbf{Regression Metrics.} To evaluate prediction performance for tasks with real-valued target variables, we employed Mean Absolute Error (MAE) and Mean Squared Error (MSE) metrics. For calculating these regression metrics, each question was accompanied by instructions specifying the format and range of the expected answer. The required numerical values (both real and integer) were then extracted from the LLM's textual predictions according to the given response structure. Instances where the prediction could not be interpreted as a number were excluded from the final metric calculation\footnote{We made this assumption based on the rarity of such instances, given the clarity of the questions and the accompanying guidance provided for answering them.}. The selected numerical responses, denoted as $y$, were compared with the target values $\hat{y}$ for accurate assessment:

\[
\label{eq:4}
\text{MAE}(y, \hat{y}) = \frac{1}{n_{\text{samples}}} \sum_{i=0}^{n_{\text{samples}}-1} \left| y_i - \hat{y}_i \right|
\]

\[
\label{eq:5}
\text{MSE}(y, \hat{y}) = \frac{1}{n_\text{samples}} \sum_{i=0}^{n_\text{samples} - 1} (y_i - \hat{y}_i)^2
\]

\subsection{Detailed datasets description} \label{B.3}

A complete list of the datasets and a description of each dataset is given below. Main statistics and descriptions for each dataset are provided in Table \ref{tab:datasets}.

\begin{table}
  \caption{Statistics of the datasets used for models evaluation.}
  \label{tab:datasets}
  \centering
  \begin{tabular}{llllll}
    \toprule
    \cmidrule(r){1-6}
    Dataset & AlfaBattle & Age & Gender & X5 & Taiwan \\
    \midrule
    \# events & 443 M. & 44 M. & 6,85 M. & 45,8 M. & 0.18 M \\
    \# sequences & 1,47 M. & 30 K. & 9,2 K. & 400 K. &  30 K.\\
    Avg, seq. len. & 881.7 & 862.4 & 446.6 & 114.3 & 6 \\
    \# numeric & 3 & 1 & 3 & 3 & 3 \\
    \# categorical & 15 & 2 & 2 & 3 & 5 \\
    \# classes & 2 & 4 & 2 & 4 & 2\\
    train/val split \% & 70/30 & 90/10 & 90/10 & 90/10 & 90/10\\
    \bottomrule
  \end{tabular}
\end{table}

\textbf{AlfaBattle2.0 dataset.} The AlfaBattle2.0 dataset \cite{Smirnov2021AlfaBattle2.0} consists of transaction activity records of bank customers over a two-year period, capturing spending, payments, and transfers. The primary goal is to estimate the probability of a customer defaulting on a loan within a given timeframe. The default rate in this dataset is 2.76\%. Each customer is associated with a sequence of transactions, each described by 18 features: 3 numeric and 15 categorical. The numeric features include the normalized transaction amount, the number of hours since the customer's last transaction, and the number of days until the loan is disbursed. The categorical features encompass various identifiers: the merchant's code and category, the currency and type of payment card, and the city, country, etc. All categories are encoded with numeric values to ensure the dataset remains anonymized. The temporal component is defined by the attributes of hour, day of the week and week of the year, which in combination form the transaction date and time. 

\textbf{Age Group Prediction Competition.} This dataset \citep{Sirius2020sberbank-sirius-lesson} comprises anonymized transaction records of bank customers, with the aim of predicting the age group of each client based on their transactions. Each transaction is characterised by three features: a discrete MCC (Merchant Category Code) identifying the type of merchant, the transaction date, and the transaction amount. Transactions can be grouped according to the unique customer identifier specified in the transaction description. The merchant identifier is also provided in text form, with categories such as 'bookshop', 'ATM', 'pharmacy', etc. This allows for a more detailed and nuanced analysis of spending patterns related to different age groups.

\textbf{Gender Prediction Competition.} The primary goal of this competition is to predict the gender of bank customers based on their transaction activity \citep{MaxValery2019python-and-analyze-data-final-project}. The dataset includes historical transaction and transfer data spanning one year and three months. Each transaction record is associated with a unique client ID and contains the time and date of the transaction, its type, the transaction amount, and a discrete identifier for the merchant point. The transaction amount is not normalised and can indicate both inflows and outflows of funds. A negative value signifies a debit, while a positive value denotes a credit to the account.

\textbf{Taiwan Default of Credit Card Clients.} This dataset \citep{Yeh2009TheCO} includes customer transaction data from April to September 2005, and it is used to predict whether a customer will repay their borrowed credit. Each record in the dataset contains 8 real-valued attributes. Some attributes describe the customer's characteristics, such as age, education level, and marital status, while the remaining attributes provide details about the history of loan repayments.

\textbf{X5 Retail Hero: Uplift Modeling for Promotional Campaign.} Initially designed for an uplift modeling competition, this dataset focuses on predicting a customer's age based on their purchasing activity \citep{babaev2022coles}. Each purchase in the dataset is characterized by the time of the transaction, product type, segment, purchase amount, and the type of loyalty program associated with the customer.

\subsection{Baselines implementation details} \label{B.4}

Below, we provide details about the architectures and hyperparameters of the baseline approaches used in our study.

\textbf{Handcrafted features with LightGBM}: This baseline aggregates numerical feature values across buckets and includes statistics such as count, mean, variance, minimum, and maximum. The LightGBM classifier \citep{Ke2017LightGBMAH} is then used for prediction.

\textbf{Randomly initialised RNN encoder}: This approach utilizes a randomly initialized and untrained RNN sequence encoder based on a unidirectional Gated Recurrent Unit (GRU) with a single hidden layer of size 1024. The resulting 1024-dimensional event sequence representations are used with LightGBM to solve the downstream task.

\textbf{CoLES} (Contrastive Learning for Event Sequences): This method employs a self-supervised contrastive pretraining approach called CoLES \citep{babaev2022coles} to generate vector representations of event sequences. The encoder is a recurrent neural network (RNN) GRU with one hidden layer of size 1024, producing final embeddings of the same size. A supervised classifier based on LightGBM is then trained using the pretrained embeddings.

\textbf{CPC} (Contrastive Predictive Coding): This approach uses a similar sequence encoder architecture to CoLES but applies the Contrastive Predictive Coding (CPC) method \citep{Oord2018RepresentationLW} for pretraining. CPC is a self-supervised technique for learning vector representations using an autoregressive model for non-discrete data sequences.

\textbf{Barlow Twins}: This method follows the same scheme and sequence encoder architecture as CoLES and CPC but implements a Barlow Twins Loss \citep{Zbontar2021BarlowTS} for encoder pre-training. LightGBM is then used on the obtained embeddings for solving the downstream problem.

\textbf{NSP} (Next Sequence Prediction): This baseline employs an RNN sequence encoder with a unidirectional GRU and a single hidden layer of size 1024, pretrained on the Next Sequence Prediction task \citep{Devlin2019BERTPO}. The resulting 1024-dimensional embeddings are used with LightGBM for the downstream task.

\textbf{RTD} (Replaced Token Detection): Similar in architecture to the NSP baseline, this approach uses the Replaced Token Detection loss function from the ELECTRA paper \citep{clark2020electra}.

\textbf{SOP} (Sequences Order Prediction): Identical in architecture to NSP and RTD, this baseline uses the Sequences Order Prediction loss function from the ALBERT work \citep{Lan2019ALBERTAL}.

\textbf{MLM NSP} (Masked Language Modelling with Next Sentence Prediction): This approach uses a LongFormer \citep{Beltagy2020LongformerTL} with 4 attention heads, 8 hidden layers of dimension 2048, and a maximum of 2000 positions as an event encoder. The output embedding size is 2048. The encoder is pretrained using a combination of Masked Language Model and Next Sentence Prediction tasks as in BERT \citep{Devlin2019BERTPO}. LightGBM is then used on the obtained embeddings for the downstream task.

\textbf{TabFormer}: This approach implements the TabFormer method \citep{Padhi2020TabularTF}, utilizing a LongFormer \citep{Beltagy2020LongformerTL} with 4 attention heads, 8 hidden layers of dimension 2048, and a maximum of 2000 positions as the sequence encoder. The output embedding size is 2048. The encoder is pretrained using the Masked Language Modelling (MLM) task \citep{Devlin2019BERTPO}. LightGBM is then used on the obtained embeddings for solving the downstream problem.

\textbf{GPT}: This approach uses a GPT-2 architecture \citep{Radford2019LanguageMA} as the event sequence encoder, with 12 layers, 12 heads per layer, and position encoding up to 2056 positions. The embedding dimension is 768. The encoder is pretrained on an autoregressive task of predicting the fields of the next transaction, each using a separate head. LightGBM is used on the obtained embeddings for the downstream task.

\textbf{RNN with CoLES}: This baseline differs from the standard CoLES approach by adding several MLP heads to the event sequence encoder after contrastive pre-training. This architecture is then end-to-end trained on the target task.

\textbf{GPT with descr.}: This approach modifies the conventional GPT-2 baseline by applying discretization to the numerical features of events.

\textbf{CatBoost}: A simple implementation of the CatBoost algorithm \citep{Ostroumova2017CatBoostUB} trained on event features.

\textbf{Text LLM}: This text-based LLM approach serializes event features into a string using a template, selecting only the attributes necessary for the task while ignoring others due to the long token sequence. The length of event sequences is also reduced to fit the language model's context. For this baseline, we used the FLAN-T5-xl \citep{Wei2021FinetunedLM} model.

%%%%%%%%%%%%%%%%%%%%%%%%%%%%%%%%%%%%%%%%%%%%%%%%%%%%%%%%%%%%

\end{document}